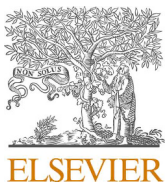
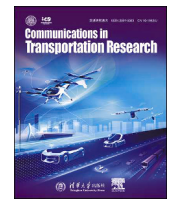

Full Length Article

# Interaction dataset of autonomous vehicles with traffic lights and signs

Zheng Li [b,d], Zhipeng Bao [c], Haoming Meng [d], Haotian Shi [a,b,*], Qianwen Li [c,**], Handong Yao [c], Xiaopeng Li [b]

[a] *College of Transportation, Tongji University, Shanghai, 201804, China*
[b] *Department of Civil and Environmental Engineering, University of Wisconsin–Madison, Madison, 53706, USA*
[c] *School of Environmental, Civil, Agricultural and Mechanical Engineering, University of Georgia, Athens, 30602, USA*
[d] *Department of Computer Science, University of Wisconsin–Madison, Madison, 53706, USA*

ABSTRACT

This study presents the development of a comprehensive dataset capturing interactions between autonomous vehicles (AVs) and traffic control devices, specifically traffic lights and stop signs. Derived from the Waymo Motion dataset, our work addresses a critical gap in the existing literature by providing real-world trajectory data on how AVs navigate these traffic control devices. We propose a methodology for identifying and extracting relevant interaction trajectory data from the Waymo Motion dataset, incorporating over 37,000 instances with traffic lights and 44,000 with stop signs. Our methodology includes defining rules to identify various interaction types, extracting trajectory data, and applying a wavelet-based denoising method to smooth the acceleration and speed profiles and eliminate anomalous values, thereby enhancing the trajectory quality. Quality assessment metrics indicate that trajectories obtained in this study have anomaly proportions in acceleration and jerk profiles reduced to near-zero levels across all interaction categories. By making this dataset publicly available, we aim to address the current gap in datasets containing AV interaction behaviors with traffic lights and signs. Based on the organized and published dataset, we can gain a more in-depth understanding of AVs' behavior when interacting with traffic lights and signs. This will facilitate research on AV integration into existing transportation infrastructures and networks, supporting the development of more accurate behavioral models and simulation tools.

## 1. Introduction

Researchers have developed various models to analyze and understand the behavior of autonomous vehicles (AVs) in different scenarios (Kuang et al., 2024), including vehicle platooning (Li and Li, 2022; Li et al., 2022a), intersection navigation (Soleimaniamiri et al., 2022), lane change maneuvers (Li et al., 2022b; Wang et al., 2021), and so on. However, these models' evaluations have relied on simulations with idealized assumptions, which often diverge significantly from real-world driving conditions. This discrepancy between the simulated and real-world environments introduces potential errors in assessing the impact of AVs on traffic dynamics and operations. To bridge this gap, it is crucial to develop calibrated driving behavior models for AVs (Huang et al., 2021). Collecting a substantial amount of trajectory data describing AV performance in real-world traffic environments is essential to developing and calibrating such models.

Recently, major organizations focusing on AV technology development have collected extensive comprehensive datasets of AV trajectories and environmental scenarios. Examples of comprehensive dataset collected by these organizations include OpenACC dataset (Makridis et al., 2021), a database derived from a series of experimental car-following campaigns designed to study commercial adaptive cruise control (ACC) systems; nuScenes dataset (Caesar et al., 2020), collected using electric vehicles in Boston and Singapore with a focus on urban road scenarios; nuPlan dataset (Caesar et al., 2021), a large-scale dataset collected from 4 cities across North America and Asia (Boston, Pittsburgh, Las Vegas, and Singapore) with widely varying traffic patterns; KITTI dataset (Geiger et al., 2013), an open-source dataset primarily used for AV perception; and Waymo Open dataset (Sun et al., 2020), composed of Waymo Motion and Waymo Perception datasets, contains high-fidelity sensor data gathered by AVs across various U.S. cities.

* Corresponding author. College of Transportation, Tongji University, Shanghai, 201804, China.
** Corresponding author.
   *E-mail addresses:* shihaotian95@tongji.edu.cn (H. Shi), cami.li@uga.edu (Q. Li).

However, most of these datasets cannot be directly used by transportation researchers for their studies. There are two main reasons for this limitation. First, these comprehensive datasets contain excessively diverse scenarios and massive volumes of data, requiring researchers to extract the specific scenario data needed for their studies from the vast collection. Second, the conversion of data formats poses a challenge, as existing datasets come in various formats that need to be processed and converted before they can be directly utilized by transportation researchers.

Addressing these needs, numerous transportation researchers have already processed these comprehensive datasets. Zhou et al. (2024) processed a unified trajectory dataset for automated vehicles' longitudinal behavior from several comprehensive datasets. Hu et al. (2022) extracted car-following segments from the Waymo Perception dataset. Similarly, Li et al. (2023) have extracted numerous car-following segments from the Lyft Level 5 dataset. The availability of these well-organized car-following data has enabled transportation researchers to conduct extensive investigations into AV car-following behavior and its impacts to current transportation system. For instance, Li et al. (2025) examined the Markov property of AV car-following behavior utilizing data from the Lyft Level 5 dataset provided by Li et al. (2023). Hu et al. (2023) employed AV car-following data extracted from the Waymo Perception dataset documented in Hu et al. (2022) to evaluate the impact of AVs on traffic flow dynamics.

Despite these advancements, current research on AV behavior understanding and dataset establishment has predominantly focused on segment-level behaviors, such as car-following maneuvers, which fails to provide a comprehensive understanding of AV operations. Comparatively, the behaviors of AV at intersections when interacting with traffic control devices (i.e., traffic light and stop sign) present more complex patterns that manifest in two aspects. First, AV behaviors are not stable and may exhibit pronounced transitions, such as sudden deceleration upon detecting red lights or stop signs, and acceleration when traffic lights turn green or after stopping at stop signs. Second, these scenarios involve intricate interactions, as AV navigating through intersections may simultaneously interact with multiple objects, including traffic lights, stop signs, leading vehicles, and potential conflict vehicles within the intersection area. These complex behaviors should have greater implications for the efficiency and safety of existing transportation systems. There is limited research examining how AV interacts with traffic signals and signs, primarily due to the absence of dedicated datasets specifically capturing these interactions. Such data deficiency creates several critical challenges: it impedes the development and calibration of driving behavior models for AV interactions with traffic signals and signs; it prevents the characterization of AV interaction behavioral patterns; and it hinders analysis of how AV interaction behaviors impact traffic flow. These limitations subsequently constrain transportation agents to establish appropriate traffic management strategies with increasing AV penetration rates (Soleimaniamiri et al., 2022). However, existing comprehensive AV datasets have already included abundant data on AV interactions with traffic lights and stop signs. Transportation researchers have not yet organized and extracted these trajectory data from comprehensive datasets to create directly useable data for transportation domain scholars.

Based on the identified gaps, this study presents the development of a dataset derived from the Waymo Motion dataset, which captures a wide range of interactions between AVs and traffic lights and signs. We have designed and implemented methods to extract trajectories that involve these interactions from the Waymo Motion Dataset. These extracted trajectory data have been further processed and refined to facilitate future research on AV behavior modeling in response to these traffic control devices. To the best of our knowledge, this represents the first dataset directly focused on AV interactions with traffic lights and signs. By making this dataset available to the research community, we aim to accelerate progress in understanding and optimizing AV behaviors in the interactions with traffic lights and signs.

The remainder of this paper is structured as follows. Section 2 reviews commonly used AV datasets. Section 3 presents our methodology for extracting AV-traffic control device interaction trajectories from the Waymo Motion dataset, along with methods for assessing and enhancing the data quality. Section 4 presents the extracted dataset and discusses the results of our quality assessment and enhancement process. Section 5 briefly introduces the approaches to use the dataset, and gives an example of using the established dataset to calibrate an AV behavior model. Finally, Section 6 concludes the study and suggests directions for future research.

## 2. Relevant dataset review

This study aims to extract and organize AV interaction data with traffic lights and signs from existing public AV datasets. There are several similar available candidates, such as OpenACC, nuScenes, nuPlan, KITTI, and Waymo Open. Three main considerations are adopted when selecting which public dataset to use for filtering AV interactions with traffic lights and stop signs. First, whether the dataset includes urban road intersection scenarios; second, whether it contains AV trajectory information that is easily accessible; and third, whether it includes traffic light status and position information, as well as stop sign position information that can be readily extracted. We utilize these requirements to evaluate the usability of the aforementioned datasets.

OpenACC dataset: OpenACC was designed to study commercial ACC systems. Data were collected on both public highways and controlled test tracks across various European locations, capturing detailed vehicle trajectory information such as speed, acceleration, and precise positioning via high-accuracy sensors and onboard systems. However, this dataset does not contain urban road intersection scenarios, as well as the traffic light information and stop sign information that we require.

nuScenes dataset: This dataset was collected using two Renault Zoe ultra-compact electric vehicles in Boston and Singapore, focusing on urban road scenarios. It used LiDAR and cameras to record AV trajectory information, surrounding vehicle trajectories, and environmental information. This dataset meets our first two requirements: it covers urban road intersection scenarios and contains easily accessible AV trajectory information. For the third requirement, it is found that while the dataset annotates the positions of traffic lights, stop lines, and stop signs, it does not directly label traffic light color information. This information would need to be extracted from camera images using computer vision methods. Therefore, while this dataset contains the traffic light and stop sign information we need, such information cannot be directly read and would require additional processing.

nuPlan: This dataset is similar to nuScenes but differs in that it developed a novel system to infer traffic light status from the motion of the vehicles. Therefore, it includes high-precision traffic light color information. However, according to its official introduction, the data was collected from human-driven vehicles (HVs), which does not meet the requirement for AV data.

KITTI: KITTI is primarily used for AV perception. The dataset was recorded using a Volkswagen station wagon, capturing approximately 6 h of traffic scenarios with various sensor modalities, including high-resolution color and grayscale stereo cameras, a Velodyne three-dimensional (3D) laser scanner, and a high-precision GPS/IMU inertial navigation system. The collected scenarios are diverse, covering real traffic situations from highways and rural roads to city interiors filled with static and dynamic objects. The dataset provides image data, vehicle GPS information, and three-dimensional point cloud data collected by the Velodyne laser scanner. This dataset meets our first two requirements: it covers urban road intersection scenarios and contains easily accessible AV trajectory information. For the third requirement, although the images and 3D point cloud data capture traffic lights and stop signs, the specific colors and positions of traffic lights and the positions of stop signs are not directly provided. Additional algorithms would be needed to extract these information from the images and point cloud data, and computer vision methods would be required to recognize traffic light colors and shapes.

Waymo Open: The Waymo Open dataset is a comprehensive collection of high-fidelity sensor data gathered by AVs across Phoenix, Mountain View, and others. A wide variety of scenarios are contained in the dataset, including car-following, lane-changing, gap acceptance, interactions with pedestrians, interactions with non-motorized vehicles, and interactions with traffic lights and signs. The dataset was collected using a sophisticated sensor array, including five LiDAR units (one mid-range and four short-range) and five cameras (covering front and side views) (Hu et al., 2022). It is found that this dataset contains numerous urban road intersection scenarios with many segments showing AV interactions with traffic lights and stop signs. Furthermore, the AV trajectory information, including position and speed, is directly provided and readily useable. Additionally, the positions, colors, and shapes of traffic lights, as well as the positions of stop signs, are also directly provided and can be used without additional processing.

The overall comparisons of these dataset are shown in Table 1. Based on above analysis, this work selects the Waymo Open dataset to extract data describing the interaction behaviors of AVs with traffic lights and signs.

## 3. Data processing methodology

### 3.1. Framework

As mentioned before, the Waymo Open dataset is categorized into two subsets: the Perception dataset and the Motion dataset. This study focuses on the Motion dataset, specifically filtering segments that include information about interactions between AVs and traffic lights or signs. Owing to Waymo's ongoing data collection and processing efforts, the number of segments in the Motion dataset continues to grow. In our analysis of V1.2.1 of the Motion dataset, we identified a total of 526,731 segments. The information in the Motion dataset includes environmental context, timestamps, AV trajectory data, background HVs trajectory data, road background information, traffic sign positions, and traffic light positions as well as their colors (i.e., red, yellow, green) and shapes (i.e., arrow, circle).

To ensure clarity and avoid ambiguity, it is important to define two key terms used to describe the data first: segment and trajectory. A segment refers to each 9.1-s clip at 0.1-s intervals of raw data extracted from the Waymo Motion dataset. Each segment contains comprehensive information (i.e., environmental context, timestamps, AV trajectory, HVs trajectories, road background information, traffic control devices states) recorded during this 9.1-s duration. A trajectory refers to the time-series data extracted and processed from a segment that can be directly used to study AV interactions with lights and signs. A trajectory includes the AV's positions, speeds, and accelerations at each time step, along with traffic light states (colors and shapes) and the positions of traffic lights and signs.

Fig. 1 illustrates the roadmap of this work in detail. First, we define rules and procedures to identify segments of AV interactions with traffic lights and signs in the Waymo Motion dataset. Then, the entire Waymo Motion dataset is traversed, and the segments that meet our defined rules are selected. The data structure of the selected segments is reorganized from the original 'tfrecord' format used in the Waymo Motion dataset into a CSV format, where each column represents a time series (i.e., the variables shown in Table 2), and each row represents a time step. In this way, the trajectories of AVs in these selected segments are extracted. Variables, such as the acceleration of the AV, the distance between the AV and traffic lights, and stop signs, are calculated. Next, we conduct an initial assessment of the obtained AV trajectories. In this work, three commonly used metrics in vehicle trajectory data evaluation are utilized to assess the quality of obtained AV trajectories, with a particular focus on identifying noise and anomalies in the trajectory data. Subsequently, we apply a trajectory denoising method to reduce the jerk and noises from the obtained AV trajectories and enhance their quality. Finally, we reassess the quality of the enhanced AV trajectories using the three same metrics.

### 3.2. Interaction with traffic light segments selection

The interaction behaviors between AV and traffic lights are divided into four categories in this work (as shown in Fig. 2): stops at traffic light, left turns at traffic light, right turns at traffic light, and straight proceeds at traffic light.

Fig. 3 presents the flowchart of rules for selecting AV and traffic light interaction segments, organized into three main decision stages. Each stage addresses a specific aspect of the interaction: whether the AV interacts with a traffic light (R1.1–R1.3), whether it stops at the stop line (R1.4–R1.6), and whether it enters the intersection (R1.7–R1.9).

The segments where interactions occur between AV and traffic lights are identified initially. A segment is considered to involve an interaction between an AV and a traffic light if the traffic light affects the AV's behavior while it navigates through the intersection. Based on this definition, we established three rules to filter the relevant segments from the Waymo Motion dataset.

**Table 1**
Summary of datasets review.

| Dataset | Whether includes urban road intersection scenarios | Whether includes AV trajectory | Whether includes directly accessible traffic light and stop sign information |
|---|---|---|---|
| OpenACC (Makridis et al., 2021) | × | ✓ | × |
| nuScene (Caesar et al., 2020) | ✓ | ✓ | × |
| nuPlan (Caesar et al., 2021) | ✓ | × | ✓ |
| KITTI (Geiger et al., 2013) | ✓ | ✓ | × |
| Waymo Open (Sun et al., 2020) | ✓ | ✓ | ✓ |

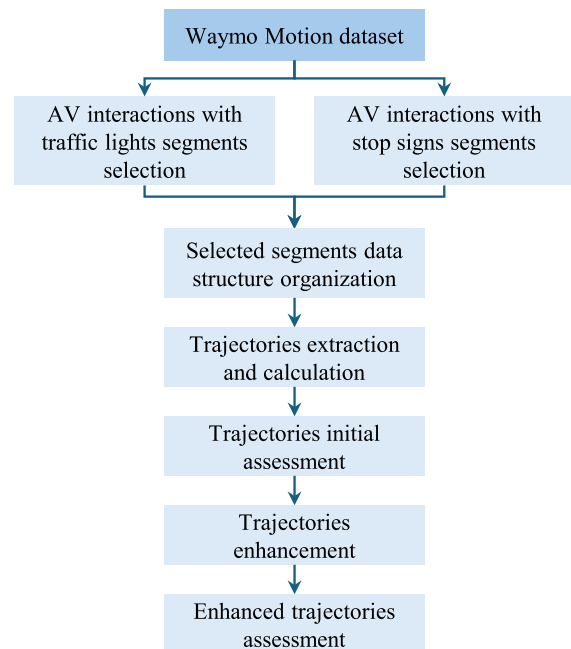

**Fig. 1.** Roadmap of the AV interaction with traffic lights and stop signs segments selection, trajectory organization, assessment and enhancement.

**Table 2**
Data structure for organized trajectories.

| Notation | Explanation |
| --- | --- |
| $P_i(x_i, y_i), \forall i \in \mathbb{N}, 1 \leq i \leq 91$ | The position coordinates of the AV at the timestep $i$, and its $x$ and $y$ components. |
| $v_i, \forall i \in \mathbb{N}, 1 \leq i \leq 91$ | The speed of the AV at the timestep $i$. |
| $a_i, \forall i \in \mathbb{N}, 1 \leq i \leq 91$ | The acceleration of the AV at the timestep $i$. |
| $L(x_l, y_l)$ | The position coordinates of the traffic light that influenced the AV, and its $x$ and $y$ components. The traffic light position in the Waymo Motion dataset is defined as the beginning of the lane in the intersection, which corresponds to the stop line at the intersection approach lane. |
| $S_\mu(x_{s,\mu}, y_{s,\mu})$ | The position coordinates of the initial nearest stop sign, which is the nearest stop sign to the AV's initial position in the segment, and its $x$ and $y$ components. |
| $u$ | The state of the traffic light that influenced the AV is coded as: unknown (0), arrow red (1), arrow yellow (2), arrow green (3), circle red (4), circle yellow (5), circle green (6), flashing red (7), and flashing yellow (8). |

R1.1 requires that the segment must include at least one traffic light. Since not all segments in the Waymo Motion dataset contain traffic lights, their presence is a fundamental criterion for determining whether a segment can be analyzed for AV-traffic light interactions.

R1.2 specifies that the AV's speed must not remain zero throughout its entire journey within the segment. If the AV's speed remains consistently zero, it indicates that the vehicle is completely stationary and the traffic light has not influenced the AV's decision-making process. In such cases, no interaction between the AV and the traffic light is considered to have occurred. Conversely, any instance of non-zero speed—whether the AV maintains a constant non-zero speed throughout the segment, decelerates to zero when approaching the traffic light, or accelerates from zero—demonstrates that the traffic light has influenced the AV's decisions and states. These scenarios are considered as interactions between the AV and the traffic light.

R1.2 is implemented through Eq. (1). To ensure that the AV is not stationary throughout the segment, we require that the cumulative time during which the speed exceeds the speed value $v_{\text{stop}}^{\text{light}}$ when the vehicle is viewed as stopped before the traffic light should be greater than $l_{\text{move}}$. $l_{\text{move}}$ is the parameter describing the minimum cumulative movement duration in the AV trajectory. $\mathbb{1}$ is the indicator function that equals 1 if the following condition is satisfied; otherwise, its value will be 0.

$$\sum_{i=1}^{91} \mathbb{1}_{v_i > v_{\text{stop}}^{\text{light}}} \geq 10 l_{\text{move}} \quad (1)$$

R1.3 requires that the AV's trajectory, or its forward extension, passes through the stop line. To implement this, we fit a $d_{\text{poly}}$th-degree polynomial to the AV's trajectory over the entire segment and determine whether this fitted trajectory intersects with the position of the stop line (as shown in Fig. 4). $d_{\text{poly}}$ is the parameter describing the degree of the polynomial to the AV's trajectory. In such cases, we consider an interaction between the AV and the traffic light to have occurred.

A special case arises when the traffic light is red, causing the AV to decelerate and stop before reaching the light. In this situation, although the AV's actual trajectory may not pass through the stop line, the traffic light has influenced the AV's behavior. To account for this scenario, we extend the fitted AV trajectory by $p_{\text{extend}}$ of its original length in the direction of travel. $p_{\text{extend}}$ is the parameter describing the forward extension percentage of the fitted AV trajectory. If the extended trajectory intersects with the stop line, this is also regarded as an interaction between the AV and the traffic light.

R1.3 is implemented through Eq. (2). The original AV positions are represented as $P_i(x_i, y_i), \forall i \in \mathbb{N}, 1 \leq i \leq 91$, and the extended positions are represented as $P_j(x_j, y_j), \forall j \in \mathbb{N}, j \geq 92$. To ensure that the AV trajectory passes through the stop line, we require that there exists at least one point on the AV's trajectory that is within $d_{\text{pass}}$ of the position of the stop line. $d_{\text{pass}}$ is the parameter describing the minimum distance threshold used to determine if an AV trajectory passes through the stop line.

$$\exists i \in \{1, \ldots, 91\}, \exists j \geq 92, \sqrt{(x_i - x_l)^2 + (y_i - y_l)^2} \\ < 0.1 \vee \sqrt{(x_j - x_l)^2 + (y_j - y_l)^2} < d_{\text{pass}} \quad (2)$$

Subsequently, we select the segments that contain interaction behaviors of the AV stopping at the stop line and the AV passing through the stop line to enter intersections.

Three rules are further devised to identify the first scenario, AV stopping at the stop line:

R1.4 is implemented through Eq. (3), requiring that the AV's speed in the first $l_{\text{begin}}$ second of the segment must exceed $v_{\text{stop}}^{\text{light}}$, indicating a significant initial speed within the beginning of the segment. $l_{\text{begin}}$ is the parameter describing the segment length threshold to split the beginning of the segment.

$$\forall i \in \left[0, 10 l_{\text{begin}}\right], v_i > v_{\text{stop}}^{\text{light}} \quad (3)$$

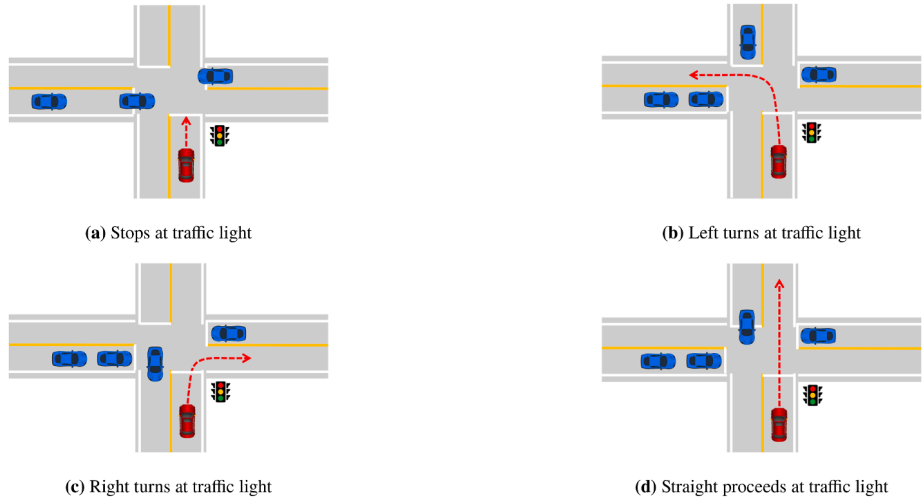

(a) Stops at traffic light

(b) Left turns at traffic light

(c) Right turns at traffic light

(d) Straight proceeds at traffic light

**Fig. 2.** Four categories of AV interactions with traffic lights.

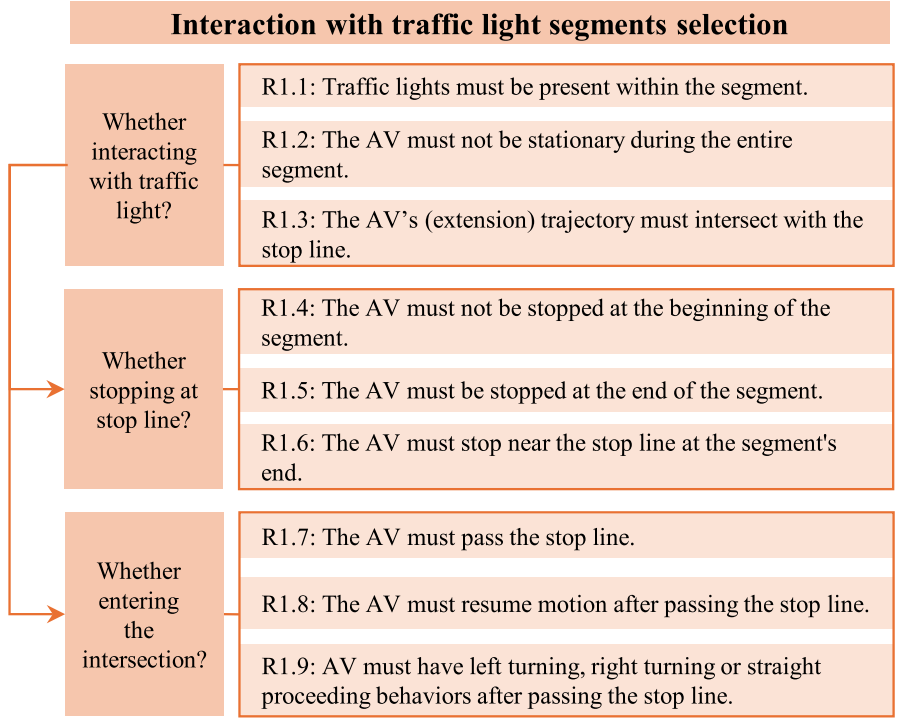

**Fig. 3.** Rules to select AV and traffic light interactions segments.

R1.5 is implemented through Eq. (4), requiring that the AV's speed in the last $l_{end}$ second of the segment be less than $v_{stop}^{light}$. $l_{end}$ is the parameter describing segment length threshold to split the ending of the segment. This rule ensures that the AV comes to a complete stop before the traffic light by the end of the segment. R1.4 and R1.5 collectively ensure that segments are captured where an AV approaches a traffic light with substantial speed and subsequently comes to a stop, indicating a clear influence of the traffic light on the AV's behavior.

$$\forall i \in [10 l_{end}, 91], v_i < v_{stop}^{light} \tag{4}$$

R1.6 stipulates that at the end of the segment, the distance between the AV and the stop line must be less than $d_{stop}$ Eq. (5). $d_{stop}$ is the parameter describing the distance threshold for determining whether the vehicle is stopped before the stop line. This criterion ensures that the AV's stopping position is directly in front of the stop line. In scenarios where multiple vehicles are queued at the intersection waiting for a red light, R1.6 guarantees that the AV is positioned at the front of the entire queue. The rationale for this rule is that if other vehicles were present in front of the AV while waiting at a red light, the behavior of the preceding vehicle would directly influence the AV's decision-making process rather than the traffic light itself. By implementing R1.6, we isolate cases where the traffic light is the primary influencer of the AV's actions.

$$\sqrt{(x_{91} - x_l)^2 + (y_{91} - y_l)^2} < d_{stop} \tag{5}$$

For the second scenario, where AVs traverse the traffic light and enter the intersection, we developed three rules to filter the relevant segments:

For R1.7, to identify segments where AV passes the stop line, it is necessary to ensure that the AV's trajectory includes movement both approaching and departing from the stop line. R1.7 is developed using two approaches, R1.7.1 and R1.7.2, specifically designed to detect this scenario. The R1.7.1 and R1.7.2 have the same purposes but approach them from different perspectives and are used together to identify relevant segments as effectively as possible.

- R1.7.1: The vectors cross product rule.
- R1.7.2: The distance rule.

The vectors cross product rule in R1.7.1 utilizes the cross product between two vectors shown in Eq. (6) to identify segments in which AV passes through the stop line (Luo and Yuan, 2012; Shaw, 1987). $\overrightarrow{P_{i-1}L}$, $\overrightarrow{LP_i}$, $\overrightarrow{P_iL}$, and $\overrightarrow{LP_{i+1}}$ represent the vectors between the AV's position and the stop line's position at different time steps. The cross products between these vectors indicate the relative positional relationship between the AV and the stop line. Equation (6) demonstrates that if there exists a time $i$ when the signs of the two vector cross products are opposite, it implies that the AV is moving towards the stop line before time $i$ and then away from it after time $i$. This change in direction suggests that the AV has passed through the stop line at approximately time $i$.

$$\forall i \in \{2, ..., 90\}, \left(\overrightarrow{P_{i-1}L} \times \overrightarrow{LP_i}\right) \times \left(\overrightarrow{P_iL} \times \overrightarrow{LP_{i+1}}\right) < 0 \tag{6}$$

Fig. 5 visualizes how the vector cross product rule is used to determine whether an AV's trajectory passes the stop line. Fig. 5a demonstrates that when the AV's trajectory passes the stop line, cross products $\overrightarrow{P_{i-1}L} \times \overrightarrow{LP_i}$ and $\overrightarrow{P_iL} \times \overrightarrow{LP_{i+1}}$ have opposite signs, resulting in a negative product $(\overrightarrow{P_{i-1}L} \times \overrightarrow{LP_i}) \times (\overrightarrow{P_iL} \times \overrightarrow{LP_{i+1}}) < 0$. Fig. 5b shows that when the AV's trajectory does not pass the stop line, cross products $\overrightarrow{P_{i-1}L} \times \overrightarrow{LP_i}$ and $\overrightarrow{P_iL} \times \overrightarrow{LP_{i+1}}$ have the same sign, resulting in a positive product $(\overrightarrow{P_{i-1}L} \times \overrightarrow{LP_i}) \times (\overrightarrow{P_iL} \times \overrightarrow{LP_{i+1}}) > 0$.

Similar to R1.7.1, R1.7.2 employs a distance-based approach to identify AV trajectories that pass through the stop line. This rule posits that if the distance between the AV and the stop line first decreases and then increases, it indicates that the AV initially approaches the stop line and subsequently moves away from it. Such a pattern of distance change suggests that the AV's trajectory has traversed the stop line.

R1.8 is implemented by requiring that the AV's trajectory extends for more than $l_{extend}$ seconds after passing through the stop line, indicating that the traffic light's influence on the AV's behavior has completely ended. $l_{\text{extend}}$ is the parameter that describes the segment length threshold to determine whether the AV travels a significant distance beyond the stop line. Additionally, this provides sufficient evidence to determine whether the AV turned left, turned right, or proceeded straight through the intersection after passing the stop line.

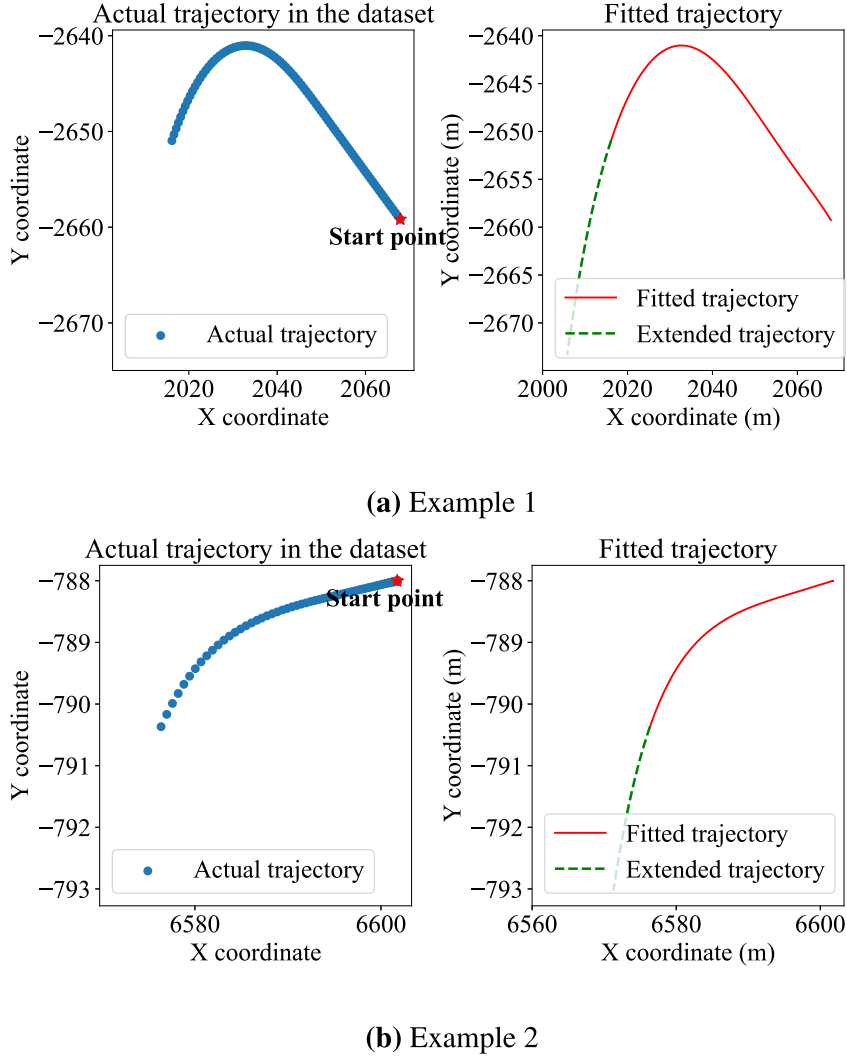

**Fig. 4.** Examples of fitting and extending the trajectory of an AV to identify whether it passes through the stop line. The original trajectory of the AV in the Waymo Motion dataset (blue points in the figure) is fitted with a polynomial (red solid line in the figure) and extended in the direction of its travel (green dashed line in the figure).

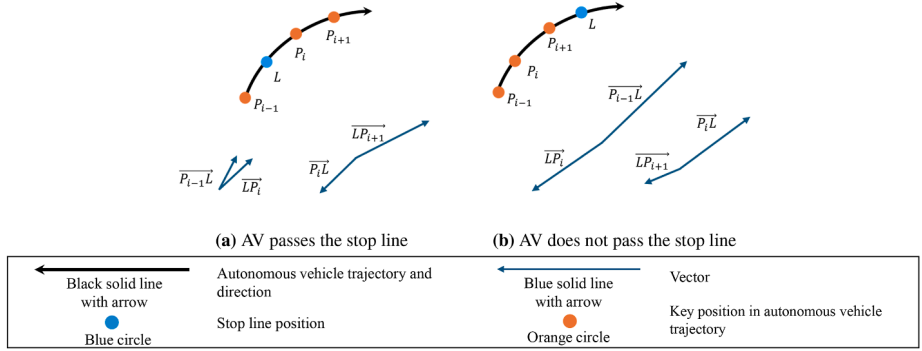

**Fig. 5.** R1.7.1 to determine whether AV passes the stop line.

R1.9 determines whether the AV turns left, right, or proceeds straight after passing the stop line using the vector cross product shown in Eq. (7) (Luo and Yuan, 2012; Shaw, 1987). In this equation, $\overrightarrow{P_1L}$ and $\overrightarrow{LP_{91}}$ are vectors connecting the AV's start and end positions to the stop line, respectively. The cross product of $\overrightarrow{P_1L}$ and $\overrightarrow{LP_{91}}$ reflects the relative positions of the AV's start, end, and the stop line. If the cross product is positive, $P_{91}$ is to the left of $\overrightarrow{P_1L}$; if negative, $P_{91}$ is to the right; if zero, $P_{91}$ is collinear with $\overrightarrow{P_1L}$. To eliminate the influence of vector length on the cross product result, we use the unit vectors $\frac{\overrightarrow{P_1L}}{|\overrightarrow{P_1L}|}$ and $\frac{\overrightarrow{LP_{91}}}{|\overrightarrow{LP_{91}}|}$ for the calculation. We stipulate that if the cross product result $\eta^{\text{light}} > \eta^{\text{light}}_{\text{left}}$, the AV turns left after passing the stop line; if $\eta^{\text{light}} < \eta^{\text{light}}_{\text{right}}$, the AV turns right; if $\eta^{\text{light}}_{\text{through},1} < \alpha < \eta^{\text{light}}_{\text{through},2}$, the AV proceeds straight. $\eta^{\text{light}}_{\text{left}}$, $\eta^{\text{light}}_{\text{right}}$, $\eta^{\text{light}}_{\text{through, 1}}$, and $\eta^{\text{light}}_{\text{through, 2}}$ are parameters describing the cross product threshold to determine whether AV turns left, turns right or proceeds straight after passing the stop line. Fig. 6 visualizes how the vector cross product is used to determine whether the AV turns left, right, or proceeds straight after passing the stop line.

$$\eta^{\text{light}} = \frac{\overrightarrow{P_1L}}{\left\|\overrightarrow{P_1L}\right\|} \times \frac{\overrightarrow{LP_{91}}}{\left\|\overrightarrow{LP_{91}}\right\|} \tag{7}$$

In summary, for any segment in the Waymo Motion dataset, sequentially evaluating rules R1.1, R1.2, R1.3, R1.4, R1.5, and R1.6 determines whether the segment contains an interaction where the AV stops before a traffic light. Similarly, evaluating rules R1.1, R1.2, R1.3, R1.7, R1.8, and R1.9 in order determines if the segment includes an interaction where the AV turns left, turns right, or proceeds straight after passing through a traffic light.

### 3.3. Interaction with stop sign segment selection

The interaction behaviors between the AV and the stop signs are divided into four categories (as shown in Fig. 7): four-way stops, right turns at the stop sign, one-step left turns at the stop sign, and two-step left turns at the stop sign.

Four-way stops: These occur at intersections where minor roads meet, with all approaches controlled by stop signs. As illustrated in Fig. 7a, vehicles from each direction, including the AV, must come to a complete stop before entering the intersection. In this scenario, the AV approaches from one of the minor roads.

Right turns at stop signs: These situations arise at junctions between major and minor roads. As shown in Fig. 7b, vehicles on the minor road, including the AV, must come to a full stop before executing a right turn, regardless of the traffic conditions on the major road. The AV then merges into the main traffic flow on the major road.

One-step left turns at stop signs: This maneuver is necessary when the major road lacks a center turning lane. As depicted in Fig. 7c, the AV must cross both directions of traffic in a single continuous movement to complete the left turn from the minor road onto the major road.

Two-step left turns at stop signs: This type of turn is executed when a center turning lane is present on the major road. As illustrated in Fig. 7d, the AV performs the turn in two distinct phases: first, it enters the center lane, and then it merges into the desired traffic flow on the major road.

Fig. 8 presents a flowchart of rules for selecting AV and stop sign interaction segments, organized into three main decision stages. Each stage addresses a specific aspect of the interaction: whether the AV is interacting with a stop sign (R2.1–R2.3), whether it's a four-way stop interaction (R2.4–R2.5), and whether the segment contains turning maneuvers at stop signs (R2.6–R2.7).

To categorize the four types of AV-stop sign interactions, we first isolate segments from the Waymo Motion dataset that exhibit such interactions. An AV-stop sign interaction segment is defined as a segment where the AV decelerates and comes to a complete stop in close proximity to a stop sign. Following three rules are developed to identify these interactions.

R2.1 requires the presence of one or more stop signs within a segment. Since not all segments in the Waymo Motion dataset include stop signs, the presence of stop signs is a basic criterion for determining whether a segment can be analyzed for AV-stop sign interactions.

R2.2 requires that the AV's trajectory must slow down when approaching the initial nearest stop sign. Since a segment may contain multiple stop signs, the stop sign closest to the AV's initial position within the segment is defined as the initial nearest stop sign, with its coordinates denoted as $S_\mu(x_{s,\mu}, y_{s,\mu})$. If the AV's speed does not decrease during this process, it is inferred that the AV might not have successfully detected the stop sign or is not traveling in a lane controlled by the stop sign. Consequently, such a segment cannot be considered as having an AV-stop sign interaction.

R2.2 is implemented using Eq. (8), which specifies that there must exist two time points, $i$ and $j$, where $i$ occurs before $j$ and satisfies two conditions. The first condition is that the distance between AV and the initial nearest stop sign decreases from time $i$ to $j$. The second condition is the AV's speed decreases from time $i$ to $j$.

$$\exists i, j \in \{1, \ldots, 91\}, i < j, \left[\sqrt{(x_i - x_{s,\mu})^2 + (y_i - y_{s,\mu})^2} > \sqrt{(x_j - x_{s,\mu})^2 + (y_j - y_{s,\mu})^2}\right] \land [v_i > v_j] \tag{8}$$

R2.3 mandates that following deceleration, the AV must come to a complete or near-complete stop at the initial nearest stop sign (Pierre and Désiré, 2022). This rule is implemented on two key aspects: the stop area and the AV's speed within this area. We designate the stop area as a circular region with a $r_{\text{stop}}$ radius, centered on the AV's location nearest to and before the initial nearest stop sign. Regarding speed, it is required that the cumulative time during which the speed is less than the value $v^{\text{sign}}_{\text{stop}}$ and the AV is within the stop area should be greater than $l_{\text{stop}}$. $l_{\text{stop}}$ is the parameter describing the minimum cumulative stopping duration in the AV trajectory.

R2.3 is implemented through Eq. (9), where $\mathbb{1}$ is the indicator function equal to 1 if the following condition is satisfied; otherwise, its value will be 0.

$$\sum_{i=1}^{91} \mathbb{1}_{v_i < v^{\text{sign}}_{\text{stop}}} \cdot \mathbb{1}_{\sqrt{(x_i - x_{s,\mu})^2 + (y_i - y_{s,\mu})^2} < r_{\text{stop}}} \geq 10 l_{\text{stop}} \tag{9}$$

Following rules are defined to select segments containing the four-

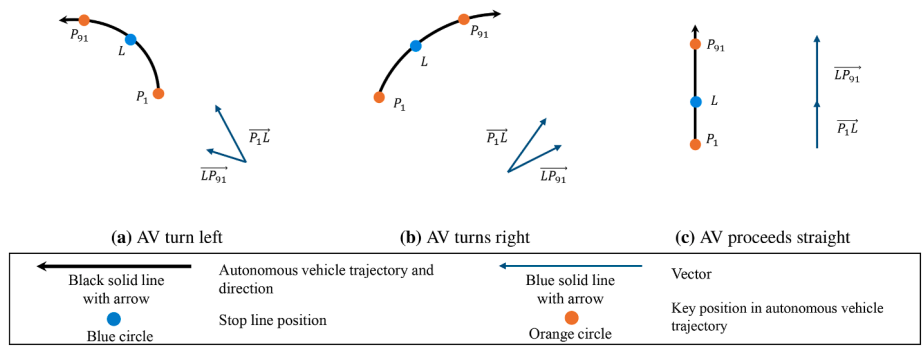

**Fig. 6.** R1.9 to determine whether the AV turns left, right, or proceeds straight after passing the stop line.

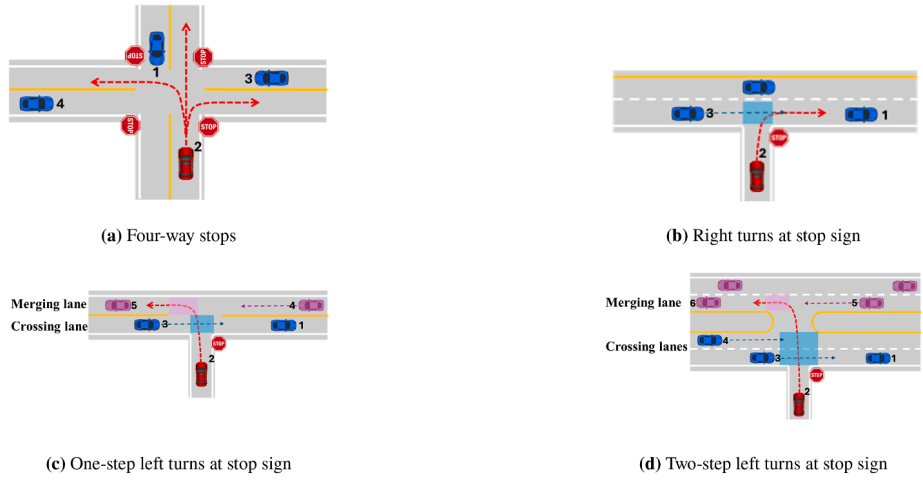

(a) Four-way stops  
(b) Right turns at stop sign  
(c) One-step left turns at stop sign  
(d) Two-step left turns at stop sign

**Fig. 7.** Four categories that AV interactions with stop sign.

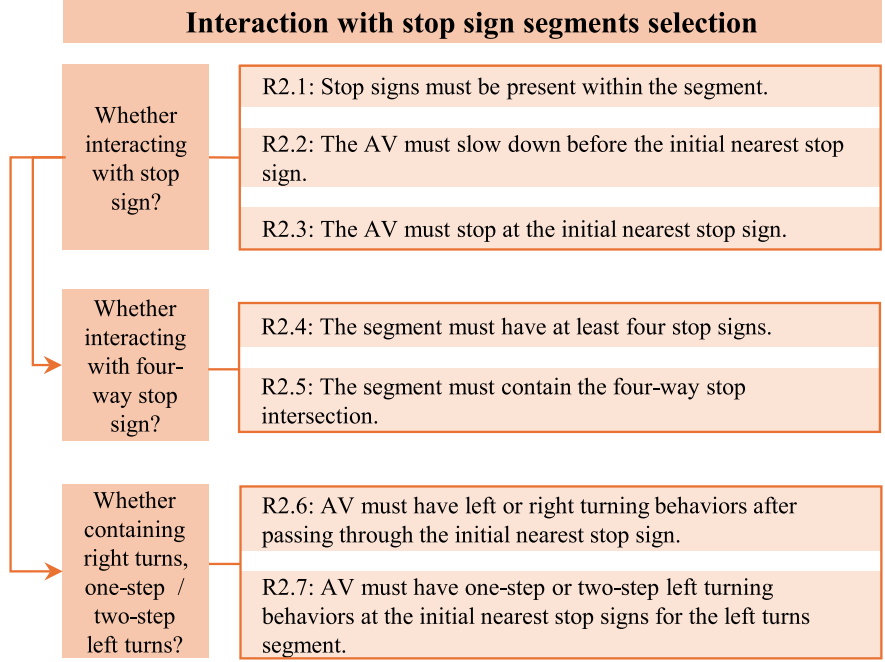

**Fig. 8.** Rules to select AV and stop sign interactions segments.

way stop AV and stop sign interaction:

R2.4 stipulates that the segment must contain four stop signs. This rule is based on the assumption that fewer than four stop signs would be insufficient to form a four-way stop interaction.

R2.5 mandates the presence of a four-way stop intersection within the segment, which is implemented by evaluating whether four stop signs configure a convex quadrilateral. Such geometric evaluation accommodates various intersection layouts beyond the standard cross shape, including rectangles, parallelograms, and trapezoids. To determine whether the stop signs form a convex quadrilateral, we employ the following procedure.

If the segment has four stop signs, other than the initial nearest stop sign $S_\mu(x_{s,\mu}, y_{s,\mu})$ defined in Table 2, three more notations are introduced to represent the remaining stop signs: $S_1(x_{s,1}, y_{s,1})$, $S_2(x_{s,2}, y_{s,2})$, $S_3(x_{s,3}, y_{s,3})$. The lowest point on the left is determined first as the reference stop sign $S_\nu(x_\nu, y_\nu)$ shown in Eq. (10):

$$\nu = \arg\min_{j\in\{1,2,3,\mu\}}\{(x_{s,j}, y_{s,j})\} \tag{10}$$

The polar angle $\theta_j (j \in \{1, 2, 3, \mu\}$, and $j \neq \nu)$ of each remaining stop sign $S_j$ relative the reference stop sign $S_\nu(x_\nu, y_\nu)$ is calculated then using Eq. (11). This angle is measured between the horizontal axis and the line connecting $S_\nu$ to $S_j$.

$$\theta_j = \arctan2(y_{s,j} - y_{s,\nu}, x_{s,j} - x_{s,\nu}), j \in \{1,2,3,\mu\}, \text{ and } j \neq \nu \tag{11}$$

Three stop signs $S_j(j \in \{1, 2, 3, \mu\}$, and $j \neq \nu)$ are sorted in ascending order of their polar angles $\theta_j$ as $S_\alpha(x_{s,\alpha}, y_{s,\alpha})$, $S_\beta(x_{s,\beta}, y_{s,\beta})$, $S_\gamma(x_{s,\gamma}, y_{s,\gamma})$. Four cross products are computed as shown in Eqs (12)–(15):

$$\eta_{\mu,\beta} = \overrightarrow{S_\mu S_\alpha} \times \overrightarrow{S_\alpha S_\beta} \tag{12}$$

$$\eta_{\alpha,\gamma} = \overrightarrow{S_\alpha S_\beta} \times \overrightarrow{S_\beta S_\gamma} \tag{13}$$

$$\eta_{\beta,\mu} = \overrightarrow{S_\beta S_\gamma} \times \overrightarrow{S_\gamma S_\mu} \quad (14)$$

$$\eta_{\gamma,\alpha} = \overrightarrow{S_\gamma S_\mu} \times \overrightarrow{S_\mu S_\alpha} \quad (15)$$

If $\eta_{\mu,\beta}, \eta_{\alpha,\gamma}, \eta_{\beta,\mu}$, and $\eta_{\gamma,\alpha}$ are uniformly positive or negative, it is concluded that these four stop signs form a convex quadrilateral, thus identifying a four-way stop intersection.

When the number of stop signs in a segment exceeds 4, it suggests the possibility of multiple four-way stop intersections or a complex intersection with additional stop signs. To identify these configurations, we employ the Density-Based Spatial Clustering of Applications with Noise (DBSCAN) (Ester et al., 1996) algorithm to group the stop signs into distinct clusters.

For each resulting cluster, we evaluate the number of stop signs. If there are exactly four stop signs in a cluster, the previously described convex quadrilateral identification procedure is applied to determine whether a four-way stop intersection can be formed. For clusters where the number of stop signs exceeds 4, DBSCAN is reapplied with the same parameters to further refine the grouping. Clusters containing fewer than four stop signs are disregarded in this analysis.

To identify scenarios involving right turns, one-step left turns, and two-step left turns at stop signs, we further establish two rules to capture the specific characteristics of each turning maneuver:

R2.6 employs a vector cross-product method (Luo and Yuan, 2012; Shaw, 1987) as formulated in Eq. (16), to determine the direction of the AV's turn at the initial nearest stop sign. This mathematical approach allows for a precise classification of right and left turns. The vectors $\overrightarrow{P_1 S_\mu}$ and $\overrightarrow{S_\mu P_{91}}$ connect the AV's start and end positions to the stop sign position, respectively. The cross product of these vectors, $\overrightarrow{P_1 S_\mu} \times \overrightarrow{S_\mu P_{91}}$, elucidates the relative spatial configuration of the AV's start position, stop sign position, and AV's end position. A positive cross product indicates that $P_{91}$ lies to the left of $\overrightarrow{P_1 S_\mu}$, while a negative result places $P_{91}$ to the right. We establish a threshold criterion for turn classification: if the normalized vector cross product $\eta^{\text{sign}} > \eta^{\text{sign}}_{\text{left}}$, we classify the maneuver as a left turn; conversely, if $\eta^{\text{sign}} < \eta^{\text{sign}}_{\text{right}}$, we categorize it as a right turn. $\eta^{\text{sign}}_{\text{left}}$ and $\eta^{\text{sign}}_{\text{right}}$ are the parameters describing the cross product threshold to determine whether the AV turns left or right after passing the stop sign.

$$\eta^{\text{sign}} = \frac{\overrightarrow{P_1 S_\mu}}{\left\|\overrightarrow{P_1 S_\mu}\right\|} \times \frac{\overrightarrow{S_\mu P_{91}}}{\left\|\overrightarrow{S_\mu P_{91}}\right\|} \quad (16)$$

R2.7 examines the AV's speed profile to distinguish between one-step and two-step left turns at stop signs. Unlike one-step left turns, AVs executing two-step left turns must come to a full stop twice: once at the initial nearest stop sign and again in the center turning lane. This behavior is reflected in the speed profile, which will show two instances where the speed falls below $v^{\text{sign}}_{\text{stop}}$, separated by a certain time interval. Consequently, we classify segments as two-step left turns when the AV's speed profile exhibits two occurrences of speeds below $v^{\text{sign}}_{\text{stop}}$ with an interval exceeding $\Delta t_{\text{stop}}$ between them. The remaining segments are categorized as one-step left turns at stop signs.

For any segment in the Waymo Motion dataset, sequentially evaluating rules R2.1, R2.2, R2.3, R2.4, and R2.5 determines whether the segment contains an AV and the four-way stop sign interaction. Similarly, assessing rules R2.1, R2.2, R2.3, R2.6, R2.7 in order determines if the segment includes an AV and stop sign interaction where the AV performs right turning, one step left turning or two turns left turning after passing the stop sign.

### 3.4. Trajectories organization

The data structure of the selected segments is converted from the original 'tfrecord' format used in the Waymo Motion dataset into a CSV format. In this format, each column represents a time series (i.e., the variables listed in Table 2), and each row corresponds to a time step. The AV trajectories are extracted from these segments, including the AV's position coordinates, speed values, and acceleration values calculated through speed differentiation. Additionally, the corresponding positions of traffic lights and stop signs are also extracted. The traffic lights state information is recorded, i.e., the color (e.g., red, yellow, and green) and the shape (e.g., arrow and circle) of the light. The distances between the AV and these traffic control devices are calculated. Consistent with the original trajectory data in the Waymo Motion dataset, our organized trajectories all have a duration of 9.1 s, with data points recorded at 0.1-s intervals.

### 3.5. Trajectories assessment metrics

The quality of the AV trajectories obtained interacting with traffic lights and signs is evaluated using three metrics: Anomaly Acceleration (%), Anomaly Jerk (%), and Anomaly Jerk Sign Inversion (%). These metrics provide a quantitative assessment of deviations from expected behavior in terms of acceleration, jerk, and jerk severity index (Li et al., 2023).

Anomaly Acceleration (%) measures the percentage of acceleration $a$ values that are considered anomalous. An acceleration value is deemed anomalous if it exceeds a predefined threshold. The normal range of acceleration is $a \in [-8 \text{ ms}^{-2}, 5 \text{ ms}^{-2}]$ (Punzo et al., 2011).

Anomaly Jerk (%) quantifies the percentage of jerk data that are identified as anomalous. Jerk ($j$) is the rate of change in acceleration over time, as shown in Eq. (17). Similar to acceleration, a jerk value is considered anomalous if it exceeds specified threshold. The normal range of acceleration is $j \in [-15 \text{ ms}^{-3}, 15 \text{ ms}^{-3}]$ (Punzo et al., 2011).

$$j = \frac{da}{dt} \quad (17)$$

Anomaly Jerk Sign Inversion measures the number of times the sign of the jerk changes within a specified time window. Frequent sign changes indicate erratic behavior or instability in motion. The jerk sign cannot be inversed more than once in 1 s (Li et al., 2023). This study documents the proportion of 1 s windows that exhibit more than one sign inversion of the jerk.

### 3.6. Trajectories enhancement methods

It has been observed that the trajectory data in the Waymo Open dataset exhibits various quality issues, including noise and outliers, despite the precision of its data collection and storage devices. These data imperfections could potentially compromise model accuracy when utilized for behavioral modeling purposes. Similar data quality concerns have been documented in other studies processing the trajectory data in the Waymo Open dataset (Hu et al., 2022). Consequently, smoothing and denoising of the raw trajectories is necessary to enhance trajectory quality.

Commonly utilized vehicle trajectory smoothing and denoising methods include discrete wavelet transform (DWT) (Fard et al., 2017), Kalman filtering (Aldimirov and Arnaudov, 2018), and spline smoothing (Cao et al., 2021). DWT decomposes signals across multiple scales, effectively capturing different frequency components and is particularly suitable for non-stationary signals with localized abrupt features, enabling noise removal while preserving essential details. Kalman filtering is appropriate when the system state transition model is known and noise follows a Gaussian distribution. Its effectiveness is highly dependent on model accuracy. The Gaussian noise assumption in Kalman filtering also requires verification. Spline smoothing represents a nonparametric curve-fitting approach that generates smooth curves by balancing data fitting precision with curve smoothness. Its fundamental principle involves identifying a smooth function that adequately fits observational data while maintaining appropriate smoothness. Spline smoothing is suitable for scenarios with gradual data variations and minimal noise.

The trajectories requiring smoothing and denoising in this study represent AV interactions with traffic lights and stop signs. During these interactions, AV may exhibit pronounced behavioral transitions, such as deceleration at red lights and stop signs, acceleration when signals turn green, and speed adjustments when encountering conflicting vehicles. A single trajectory may encompass entirely different behaviors, such as stopping at a stop sign followed by accelerating through an intersection. Given these complex and transitional characteristics of interaction trajectories, spline smoothing methods suitable for gradual trajectories may exhibit over-smoothing problems when capturing localized rapid changes. For Kalman filtering, selecting an appropriate system state transition model presents difficulties, and the Gaussian noise distribution assumption may not be applicable to the trajectories in the Waymo Motion dataset. Consequently, this work employs the DWT method for trajectory smoothing and denoising. Previous research has also utilized DWT for vehicle trajectory smoothing, including Fard et al. (2017), which applied this method to denoise and smooth NGSIM trajectory data, and Hu et al. (2022), which smoothed car-following scenario trajectories from the Waymo Perception dataset.

The DWT-based trajectory smoothing and denoising begins with the selection of an appropriate wavelet type. For our purposes, the Daubechies 6 (db6) wavelet is chosen after trial and error, due to its capability to handle data with sharp transitions—a common characteristic in vehicle dynamics. The db6 wavelet provides a good balance between data smoothness and the ability to preserve essential features, such as edges and peaks in the speed profile.

Using the selected wavelet, the noisy trajectory data is decomposed into multiple levels of detail and approximation coefficients through wavelet decomposition. This decomposition splits the signal into a series of finer-scale detail coefficients that capture high-frequency components (generally noise and fine details) and approximation coefficients that represent the low-frequency components (the underlying trends).

After decomposition, a naïve thresholding approach is applied, where all detail coefficients at each level of decomposition are set to zero. This approach is based on the assumption that for the short-duration and high-resolution data typical of vehicle trajectories, most of the significant noise can be attributed to these high-frequency components. By setting these coefficients to zero, we remove a substantial portion of the noise, simplifying the signal while retaining the overall structural integrity of the vehicle's movement.

Finally, the denoised trajectory is reconstructed using the inverse wavelet transform. This step uses the modified coefficients, now devoid of the smaller-scale noise components, to synthesize a cleaner version of the original signal. The approximation coefficients ensure that the essential characteristics of the original trajectory are preserved.

## 4. Interaction dataset

### 4.1. Interactions with traffic lights

The rules established in Section 3.2 are used to select segments containing traffic light and AV interactions. The detailed parameter values adopted during the selection of the AV and traffic light interaction segment are shown in Table 3. The following explains the rationale for selecting these parameter values.

- $l_{move}$: The requirement that the AV must have at least 1 s of non-stationary movement ensures the trajectory captures dynamic vehicle behavior. This 1-s threshold is determined empirically.
- $v_{stop}^{light}$: Since the Waymo dataset does not contain absolute zero speed values, speeds below 1 m/s are established as the threshold for stationary status.

**Table 3**
Detailed parameter values used to select segments containing AV and traffic light interactions.

| Notation | Explanation | Value |
| --- | --- | --- |
| $l_{move}$ | Minimum cumulative movement duration in the AV trajectory | 1 s |
| $d_{pass}$ | Minimum distance threshold used to determine if an AV trajectory passes through the stop line | 0.1 m |
| $d_{poly}$ | Degree of the polynomial to the AV's trajectory | 6 |
| $p_{extend}$ | Forward extension of the fitted AV trajectory | 20% |
| $v_{stop}^{light}$ | Speed threshold for determining whether the vehicle is stopped when interacting with the traffic light | 1 m/s |
| $l_{begin}$ | Segment length threshold to split the beginning of the segment | 1 s |
| $l_{end}$ | Segment length threshold to split the ending of the segment | 1 s |
| $d_{stop}$ | Distance threshold for determining whether the vehicle is stopped before the stop line | 5 m |
| $l_{extend}$ | Segment length threshold to determine whether AV travels a significant distance beyond the stop line | 2 s |
| $\eta_{left}^{light}$ | Cross product threshold to determine whether AV turns left after passing the stop line | 0.3 |
| $\eta_{right}^{light}$ | Cross product threshold to determine whether AV turns right after passing the stop line | −0.3 |
| $\eta_{through,1}^{light}$, $\eta_{through,2}^{light}$ | Cross product thresholds to determine whether AV proceeds straight after passing the stop line | 0.1, −0.1 |

- $d_{ploy}$, $p_{extend}$: Through iterative testing, a 6th-degree polynomial with a 20% trajectory extension is determined to optimally represent the majority of trajectories in the Waymo Motion dataset.
- $d_{pass}$: Due to inherent precision limitations in coordinate data, intersection points are defined as those with distances less than 0.1 m from each other.
- $l_{begin}$, $l_{end}$: These parameters designate the first and last seconds of the trajectory as the start and end times respectively.
- $d_{stop}$: A 5-m threshold for the distance between the AV and stop line, approximately equivalent to one vehicle length, ensures no intermediate vehicles between the AV and the stop line.
- $l_{extend}$: A 2-s parameter is established to preserve information about the AV's continued movement after crossing the stop line, based on empirical observation.
- $\eta_{left}^{light}$, $\eta_{right}^{light}$: While theoretically positive values indicate left turns and negative values indicate right turns, thresholds of 0.3 and −0.3 are implemented to account for natural roadway curvature at intersections that might affect trajectory direction.
- $\eta_{through,1}^{light}$, $\eta_{through,2}^{light}$: Theoretical straight trajectories would yield zero values, but practical calculations and intersection geometries necessitated thresholds of 0.1 and −0.1 for straight movement classification.

All segments in the Waymo Motion dataset are iteratively examined. Segments that satisfy the established rules are recorded. The data structure of these selected trajectory segments is reorganized. The trajectories of AV and the position, as well as the state information of the related traffic light in these selected segments, are extracted. Then, the quality of the trajectory of AV is evaluated and enhanced.

It should be clarified that most parameter values are empirically selected based on our observations in the Waymo Motion dataset. The other parameter values are referenced from the design characteristics of intersections in the United States. We acknowledge that changes in parameters might lead to variations in the final organized trajectories. The current parameter values may not necessarily be optimal. Although we traversed the entire Waymo Motion dataset using the current parameter values and rules, we may not have filtered out all instances of AV interactions with traffic lights in this section and stop signs in the following section.

Table 4 provides a summary of AV and the traffic light interaction

**Table 4**
Summary of AV and traffic light interaction trajectories.

| Category | Segments quantity | Distance (km) | Duration (h) | Anomaly acceleration proportion (%) | Anomaly jerk proportion (%) | Anomaly jerk sign inversion proportion (%) |
| --- | --- | --- | --- | --- | --- | --- |
| Stops at traffic light | 13,397 | 245.77 | 33.86 | 0.01/0.00 | 0.22/0.00 | 99.16/73.44 |
| Left turns at traffic light | 4730 | 206.19 | 11.96 | 0.23/0.00 | 9.26/0.00 | 98.23/63.91 |
| Right turns at traffic light | 3071 | 140.26 | 7.76 | 0.24/0.00 | 8.72/0.00 | 98.48/73.44 |
| Straight proceeds at traffic light | 16,379 | 1321.73 | 41.40 | 0.17/0.00 | 0.73/0.00 | 97.94/63.44 |

**Note:** The numbers before and after the slash represent the trajectory data quality assessment results before and after enhancement, respectively. For example, "0.01/0.00" means that the trajectory data quality assessment result is 0.01 before enhancement and 0.00 after enhancement.

trajectories organized in this study. The trajectories are classified into four categories: stops at traffic light (13,397 trajectories), left turns at traffic light (4730 trajectories), right turns at traffic light (3071 trajectories), and straight proceeds at traffic light (16,379 trajectories). Fig. 9 presents four categories of AV-traffic light interactions. The straight proceeds category represents the largest portion of the data, covering 1321.73 km over 41.40 h, followed by the stops category with 245.77 km over 33.86 h. The quality assessment of these trajectories before enhancement reveals that turning maneuvers (both left and right) exhibit higher anomalies in acceleration (around 0.23%) and jerk (8%–9%) compared to stopping and proceeding straight scenarios. The Anomaly Jerk Sign Inversion proportions are notably high (97%–99%) across all categories before enhancement. After applying the enhancement process, both Anomaly Acceleration and Jerk proportions are reduced to 0% across all categories, while the Anomaly Jerk Sign Inversion proportions show significant improvement, decreasing to around 63%–73%, indicating substantially smoother trajectories suitable for further analysis.

Fig. 10 demonstrates the effectiveness of the wavelet-based denoising approach through two examples. Each example shows the speed (upper) and acceleration (lower) profiles, where red dashed lines represent original data and green solid lines show denoised results. In both cases, the denoising approach effectively eliminates anomalies while maintaining the essential characteristics of the AV's motion.

### 4.2. Interactions with stop signs

The rules established in Section 3.3 are utilized to select segments containing AV and stop-sign interactions. The detailed parameter values adopted during the selection of the AV and stop sign interaction segment are shown in Table 5. The following explains our rationale for selecting these parameter values.

- $r_{stop}$: The circular stop area radius is established considering standard urban lane widths in USA.
- $v_{stop}^{sign}$: The reason is similar to the selection the value of $v_{stop}^{light}$ in AV and traffic light interaction.
- $l_{stop}$, $\Delta t_{stop}$: These parameters are derived from Pierre and Désiré (2022) defining vehicle stopping duration parameters at stop signs.
- $\eta_{left}^{sign}$, $\eta_{right}^{sign}$: The reason is similar to the selection the values of $\eta_{left}^{light}$, $\eta_{right}^{light}$ in AV and traffic light interaction.

The DBSCAN parameters are set as follows in R2.5: the neighborhood radius is set to 28 m, and the minimum number of points is 2. These choices are guided by the spatial characteristics of urban four-way stop

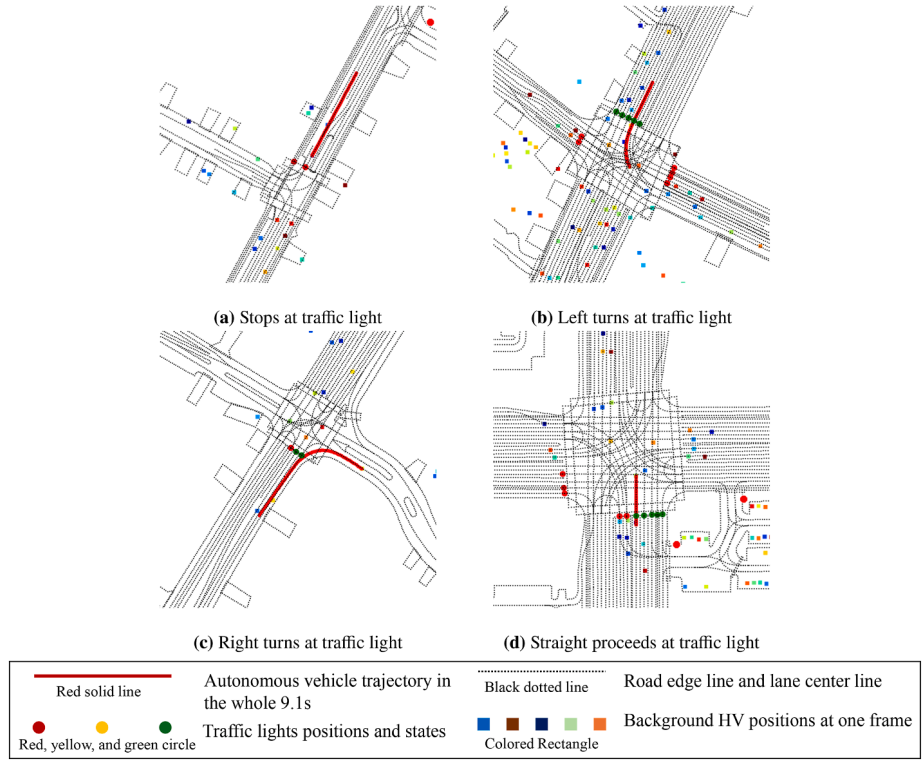

**Fig. 9.** Four different categories of AV and traffic light interactions found in the Waymo Motion dataset. The red line represents the AV trajectory, the colored circles (red, yellow, and green) represent the traffic lights and their corresponding states, and the colored rectangles represent background HVs.

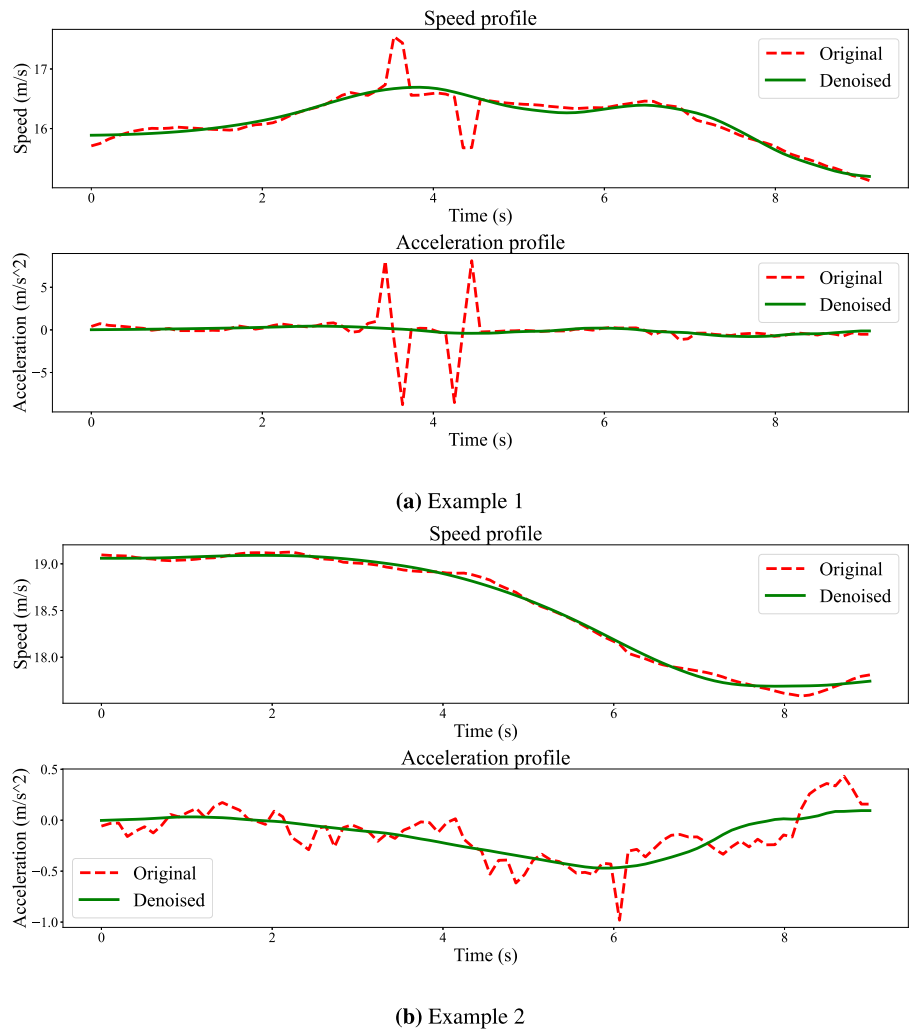

**(a)** Example 1

**(b)** Example 2

**Fig. 10.** Two examples illustrating the differences in AV trajectories before and after applying the wavelet-based denoising approach.

**Table 5**
Detailed parameter values used to select segments containing AV and stop sign interactions.

| Notation | Explanation | Value |
| --- | --- | --- |
| $r_{stop}$ | Radius for the circular stop area | 5 m |
| $l_{stop}$ | Minimum cumulative stopping duration in the AV trajectory | 0.5 s |
| $\eta_{left}^{sign}$ | Cross product threshold to determine whether AV turns left after passing the stop sign | 0.3 |
| $\eta_{right}^{sign}$ | Cross product threshold to determine whether AV turns right after passing the stop sign | −0.3 |
| $v_{stop}^{sign}$ | Maximum speed for determining whether the vehicle is stopped when interacting with the stop sign | 0.5 m/s |
| $\Delta t_{stop}$ | Time interval threshold for distinguishing between one-step left turn and two-step left turn | 1 s |

intersections, as well as best practices for DBSCAN parameter selection discussed in Schubert and Sester, 2017. In particular, the neighborhood radius is chosen based on the typical scale of such intersections to group nearby stop signs while avoiding the merging of distinct intersections. Given that standard lane widths range from approximately 3 to 3.7 m, and a four-way stop intersection typically spans from 20 to 40 m, the neighborhood radius value of 28 m ensures that stop signs within a reasonable proximity are grouped together, while still preventing the merging of distinct intersections. The minimum number of points is set to 2 to enhance the stability of the clustering process. A single stop sign alone does not define an intersection, so requiring at least two stop signs within the neighborhood radius prevents small, isolated stop signs from forming spurious clusters.

We iteratively examine the Waymo Motion dataset, select segments meeting our established rules, extract AV trajectories and stop sign information, and enhance the trajectory quality through our processing pipeline.

Table 6 provides a summary of AV and stop sign interaction trajectories organized in this study. Trajectories are classified into four categories: four-way stops (29,682 trajectories), right turns at stop sign (9670 trajectories), one-step left turns at stop sign (5189 trajectories), and two-step left turns at stop sign (14 trajectories). Fig. 11 presents these four categories of AV-stop sign interactions. The four-way stop category represents the largest portion of the data, covering 580.67 km over 74.20 h, followed by the right turn category with 367.93 km over 24.17 h. The quality assessment of these trajectories before enhancement reveals that turning maneuvers exhibit higher anomalies in acceleration (around 0.16%–0.17%) and jerk (5%–9%) compared to four-

**Table 6**
Summary of AV and stop sign interaction trajectories.

| Category | Segments quantity | Distance (km) | Duration (h) | Anomaly acceleration proportion (%) | Anomaly jerk proportion (%) | Anomaly jerk sign inversion proportion (%) |
| --- | --- | --- | --- | --- | --- | --- |
| Four-way stops | 29,682 | 580.67 | 74.20 | 0.06/0.00 | 2.79/0.00 | 98.60/59.62 |
| Right turns at stop sign | 9670 | 367.93 | 24.17 | 0.17/0.00 | 5.76/0.00 | 98.71/59.58 |
| One-step left turns at stop sign | 5189 | 188.74 | 12.97 | 0.16/0.00 | 9.05/0.00 | 98.65/61.03 |
| Two-step left turns at stop sign | 14 | 0.43 | 0.04 | 0.17/0.00 | 7.34/0.00 | 96.52/61.65 |

**Note:** The numbers before and after the slash represent the trajectory data quality assessment results before and after enhancement, respectively. For example, "0.06/0.00" means that the trajectory data quality assessment result is 0.06 before enhancement and 0.00 after enhancement.

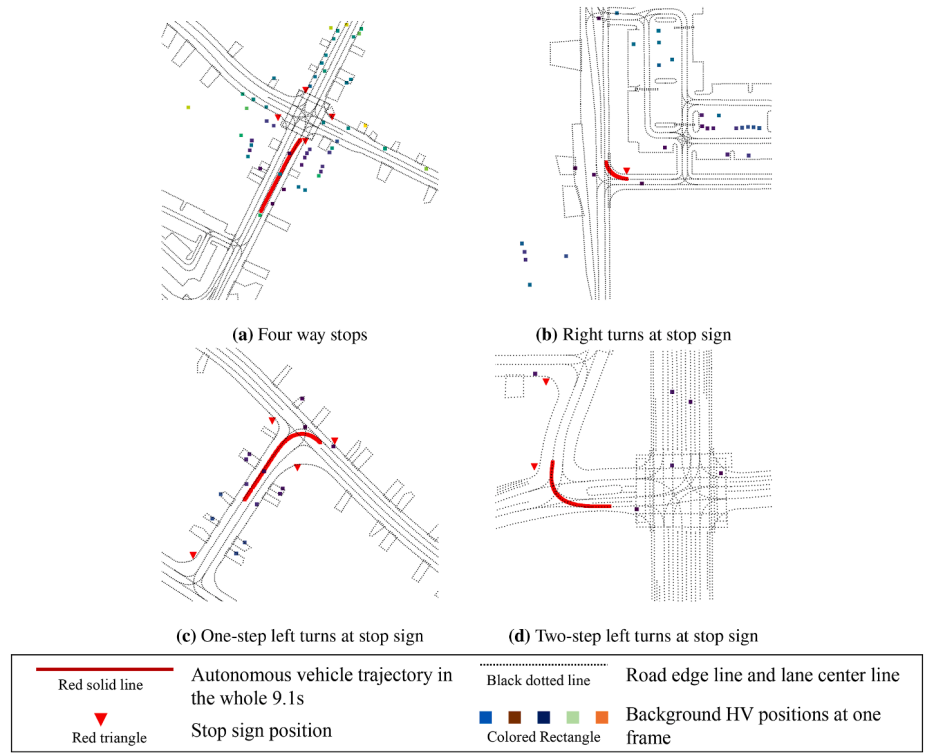

**Fig. 11.** Four different categories of AV and stop sign interactions found in the Waymo Motion dataset. The red line represents the AV trajectory, the red triangles are stop signs, and the colored rectangles represent background HVs.

**Table 7**
IDM calibration results.

| $v_0$ | $T$ | $a_{max}$ | $b$ | $s_0$ | $\delta$ | RMSE on calibration trajectories | RMSE on validation trajectories |
| --- | --- | --- | --- | --- | --- | --- | --- |
| 10.11 | 2.17 | 0.25 | 2.31 | 4.83 | 4.96 | 0.3633 | 0.3257 |

way stop scenarios. The Anomaly Jerk Sign Inversion proportions are notably high (96%–98%) across all categories before enhancement. After applying the enhancement process, the Anomaly Acceleration and Jerk proportions are reduced to 0% in all categories, while the Anomaly Jerk Sign Inversion proportions show a significant improvement, decreasing to around 59%–61%, indicating substantially smoother trajectories suitable for further analysis.

## 5. Dataset application

### 5.1. Application description

This work establishes a dataset of AV interactions with traffic lights and stop signs based on the Waymo Motion dataset. The established dataset has the following potential applications.

- AV behavior modeling at traffic control devices: Researchers can use this dataset to study and model AV behavior characteristics near traffic lights and stop signs, particularly for decision-making processes and motion patterns.
- Traffic simulation enhancement: The dataset offers empirical data to calibrate and validate traffic simulation models that incorporate AVs, improving the realism of simulated AV and traffic control devices interactions.
- Comparative studies: Researchers can analyze differences between human driver behavior and AV behavior at intersections to identify potential improvements in traffic flow efficiency.
- Algorithm development and validation: Machine Learning and rule-based algorithms for AV navigation at intersections can be developed and benchmarked using dataset.

To use the data provided in this work, researchers should first identify the specific interaction type relevant to their study from our categorization (shown in Figs. 2 and 7). While we provide smoothed data, users may apply additional filtering based on their specific requirements. Our methodological framework also allows for extension to other public AV datasets following similar principles.

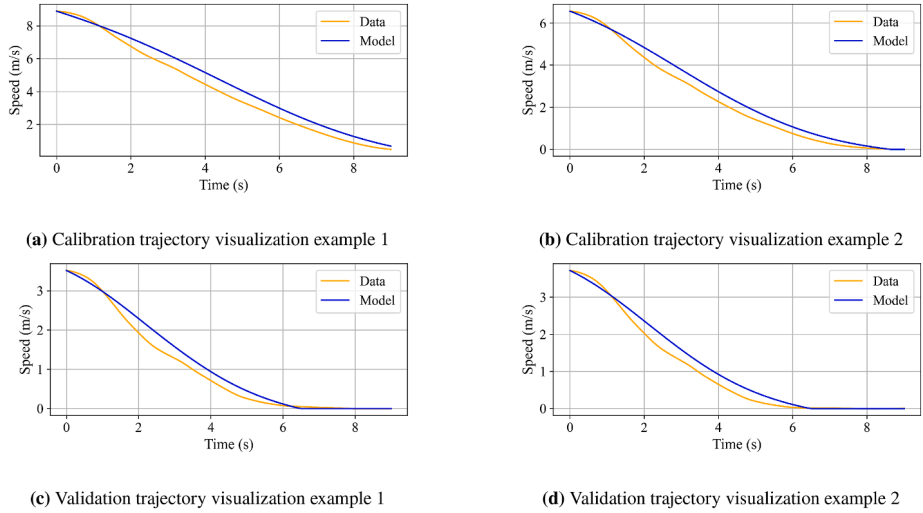

**Fig. 12.** Comparisons of Waymo trajectories and calibrated trajectories.

### 5.2. Application example

This section demonstrates the practical application of the established dataset through a simple example. We have selected a fundamental interaction between an AV and a traffic light, the behavior of an AV stopping at a red light, to illustrate the dataset's utility. The following discussion presents the complete process of developing a behavioral model for AVs at red light intersections using the proposed dataset.

19 trajectories are selected from the "Stops at traffic light" category of AV-traffic light interactions. Of these, 15 trajectories are allocated for model calibration, while the remaining 4 are reserved for model validation. As detailed in Table 2, each selected trajectory contains time-series information including the AV's position, speed, acceleration, the position of the influencing traffic light (which corresponds to the stop line position in the Waymo dataset), and the state of the influencing traffic light.

For characterizing the AV stopping behavior at red lights, we employ the intelligent driver model (IDM) to model vehicle deceleration patterns when approaching a red light. The formulation of the IDM is shown as Eqs. (18) and (19), where $a$ is the acceleration of the AV, $a_{max}$ is the maximum acceleration, $v$ is the speed of the AV, $v_0$ is the desired speed, $\delta$ is the acceleration exponent, $s$ is the distance between AV and the stop line, $s^*(v, \Delta v)$ is the desired minimum gap, $s_0$ is the minimum spacing, $T$ is the desired time headway, $b$ is the comfortable braking deceleration. In Eqs. (18) and (19), $a$ $v$, and $s$ are derived from the trajectory data, while $a_{max}$, $v_0$, $\delta$, $s_0$, $T$, and $b$ are parameters that require calibration.

$$a = a_{max}\left[1 - \left(\frac{v}{v_0}\right)^\delta - \left(\frac{s^*(v, \Delta v)}{s}\right)^2\right] \quad (18)$$

$$s^*(v, \Delta v) = s_0 + vT + \frac{v^2}{2\sqrt{a_{max}b}} \quad (19)$$

The Monte Carlo sampling methodology are employed to determine the optimal parameter values for $a_{max}$, $v_0$, $\delta$, $s_0$, $T$, and $b$ with 15 trajectories designated for calibration. The calibrated parameters are then validated against separate validation trajectories. The root mean square error (RMSE) between the acceleration values from the trajectory data and those calculated by the IDM is used as the evaluation metric for the calibration results. The calibrated parameters and evaluation results are presented in Table 7. Fig. 12 illustrates the visual comparison between the calibrated IDM and the original speed trajectories, where Figs. 12a and 12b shows examples from two calibration trajectories and Figs. 12c and 12d presents examples from two validation trajectories.

Based on the combined evidence from Table 7 and Fig. 12, it could be concluded that the established IDM accurately fits the AV-traffic light interaction trajectory data established in this research, particularly in precisely describing AV stopping behavior at traffic lights. From another perspective, this confirms that the constructed AV-traffic light interaction trajectory dataset is reasonable and applicable. These trajectories effectively captures the behavioral characteristics of AVs when interacting with traffic lights, and can be employed for studying and modeling AV behavior during traffic control devices interactions.

### 6. Conclusions

This study presents the development of a dataset that captures interactions between AVs and traffic control devices, specifically traffic lights and stop signs. Derived from the Waymo Motion Dataset, our work addresses a critical gap in the existing literature by providing real-world data on how AVs navigate these crucial elements of traffic infrastructure.

Key contributions of this work include.

1) The establishment of a systematic methodology to identify and extract relevant AV interaction segments from an existing AV dataset.
2) The development of detailed classification rules for various types of AV interactions with traffic lights and stop signs, including stopping, turning, and proceeding through intersections.
3) The creation of a high-quality dataset that can support further research into AV behavior at intersections with control devices.

The resulting dataset, which includes over 37,000 segments of AV interactions with traffic lights and more than 44,000 segments of interactions with stop signs, provides a rich resource for the research community. This data can be instrumental in developing more accurate models of AV decision-making processes when facing these traffic control devices in the intersections, further improving traffic simulation tools, and informing the design of future intelligent transportation systems.

Future work could extend this approach to include interactions with other types of traffic control devices, such as yield signs and speed limit signs, as well as more complex traffic scenarios, such as simultaneously interacting with traffic control devices and other HVs, pedestrians, and bicycles. Additionally, the methodology developed in this work could be applied to other large-scale AV datasets to create a more comprehensive understanding of AV interaction behaviors with traffic control devices. Additionally, it's important to acknowledge that our established dataset currently only contains AV trajectory, traffic light and stop sign information. It should be recognized that intersections always involve numerous traffic participants. Vehicles from other directions may influence AV behavior, especially when the AV does not have the highest right-of-way. We will continue to refine this work in the future to include the information about other intersection participants. Any future updates and maintenance to the dataset will be promptly published through our provided dataset address.

**CRediT authorship contribution statement**

**Zheng Li:** Conceptualization, Writing – original draft, Data curation, Writing – review & editing, Methodology. **Zhipeng Bao:** Writing – original draft, Data curation, Methodology. **Haoming Meng:** Data curation, Software. **Haotian Shi:** Supervision, Writing – original draft, Writing – review & editing. **Qianwen Li:** Project administration, Supervision, Funding acquisition. **Handong Yao:** Funding acquisition, Project administration. **Xiaopeng Li:** Project administration, Funding acquisition, Supervision.

**Replication and data sharing**

The interaction dataset established in this work is available at ETSData (https://doi.org/10.26599/ETSD.2025.9190035) and GitHub (https://github.com/CATS-Lab/Data-AV-Traffic-Light-Stop-Sign).

**Declaration of competing interest**

The authors declare that the research was conducted in the absence of any commercial or financial relationships that could be construed as a potential conflict of interest. While this research utilizes the Waymo Motion dataset, none of the authors are affiliated with or have received direct funding from Waymo. The dataset analysis and processing were conducted independently. This work was supported by the Federal Highway Administration (FHWA) under Broad Agency Announcement (BAA) Award (Number 693JJ324C000003), and computing resources were provided by the Center for High Throughput Computing at the University of Wisconsin–Madison. The funders had no role in study design, data collection and analysis, decision to publish, or preparation of the manuscript.

**Acknowledgements**

This work was supported by the Federal Highway Administration (FHWA) under Broad Agency Announcement (BAA) Award Number (Grant No. 693JJ324C000003). We gratefully acknowledge the support provided by the FHWA. Additionally, we acknowledge the support from the Center for High Throughput Computing (2006) at the University of Wisconsin–Madison for data processing and storage.

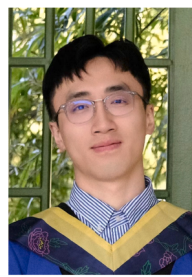

**Zheng Li** received the B.S. degree in civil engineering from Hunan University, China, in 2020 and the M.S. degree in transportation engineering from Tongji University, China, in 2023. He is currently a Ph.D. Student and Research Assistant with the Department of Civil and Environmental Engineering, University of Wisconsin–Madison. His major research interests include autonomous vehicle control, evaluation, and simulation-based optimization for traffic systems design.

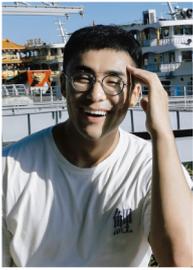

**Zhipeng Bao** received the B.S. degree in automotive engineering from Wuhan University of Technology, and the M.S. degree in automotive engineering from Jilin University, China. He is currently pursuing the Ph.D. degree in mechanical engineering at the University of Georgia, USA. His research interests lie in embodied AI, autonomous driving integrated with large language models. He has worked on vehicle trajectory prediction and SLAM systems.

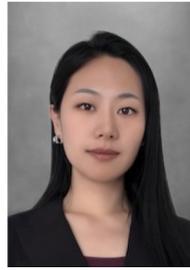

**Qianwen Li** received the B.S. degree in computer science and technology from Shandong University, China, in 2018, and the M.S. and Ph.D. degrees in transportation engineering from the University of South Florida, in 2020 and 2022, respectively. She is an Assistant Professor with the School of Environmental, Civil, Agricultural and Mechanical Engineering, University of Georgia. Her main research interests are intelligent transportation systems and transportation safety.

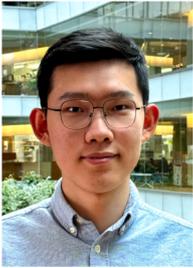

**Haoming Meng** received the B.S. degree in computer science from the University of Wisconsin–Madison in 2023. He is currently a Research Software Engineer at the Center for High Throughput Computing (CHTC) at University of Wisconsin–Madison. His major interests include high-throughput computing and distributed data storage systems. He is also a contributor to Pelican, an open-source platform for federating scientific data.

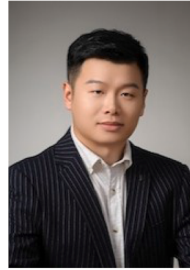

**Handong Yao** is an Assistant Professor in the College of Engineering at the University of Georgia. His research is mainly focused on developing cyber-physical transportation systems with smart infrastructure technologies and AI techniques and using machine learning methods and traffic flow theory to improve system sustainability, safety, and efficiency.

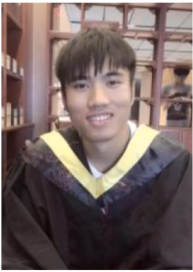

**Haotian Shi** is a tenured Associate Professor at Tongji University. He received the Ph.D. degree in civil and environmental engineering from the University of Wisconsin–Madison in 2023. He also received three M.S. degrees in power and machinery engineering (Tianjin University, 2020), civil and environmental engineering (University of Wisconsin–Madison, 2020), and computer sciences (University of Wisconsin–Madison, 2022). His main research directions are connected and automated vehicles, intelligent transportation systems, and Artificial Intelligence.

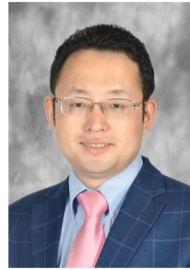

**Xiaopeng Li** received the B.S. degree in civil engineering with a computer engineering minor from Tsinghua University, China, in 2006, and the dual M.S. degrees in civil engineering and in applied mathematics and the Ph.D. degree in civil engineering from the University of Illinois at Urban–Champaign, USA, in 2007, 2010, and 2011, respectively. He is currently a Professor with the Department of Civil and Environmental Engineering, University of Wisconsin–Madison. He has published around 70 peer-reviewed journal articles. His major research interests include automated vehicle traffic control and connected and interdependent infrastructure systems.